\documentclass[letterpaper]{article} 
\usepackage{aaai2026}  
\usepackage{booktabs}
\usepackage{multirow}
\usepackage{times}  
\usepackage{helvet}  
\usepackage{courier}  
\usepackage[hyphens]{url}  
\usepackage{xcolor}
\usepackage{graphicx} 
\usepackage{amsmath, amssymb}
\usepackage{comment}
\usepackage{natbib}  
\usepackage{caption} 
\usepackage{algorithm}
\usepackage{algorithmic}

\usepackage{adjustbox}

\usepackage{newfloat}
\usepackage{listings}
\DeclareCaptionStyle{ruled}{labelfont=normalfont,labelsep=colon,strut=off} 
\floatstyle{ruled}
\newfloat{listing}{tb}{lst}{}
\floatname{listing}{Listing}
\title{Time-Series at the Edge: Tiny Separable CNNs for Wearable Gait Detection and Optimal Sensor Placement}

\author{
    Andrea Procopio\thanks{Corresponding author: aprocopio@fas.harvard.edu}\textsuperscript{\rm 1,4},
    Marco Esposito\textsuperscript{\rm 2},
    Sara Raggiunto\textsuperscript{\rm 2},\\
    Andrey Gizdov\textsuperscript{\rm 1,3},
    Alberto Belli\textsuperscript{\rm 2},
    Paola Pierleoni\textsuperscript{\rm 2}
}

\affiliations{
    \textsuperscript{\rm 1}Harvard University, Cambridge, MA, USA \quad
    \textsuperscript{\rm 2}Polytechnic University of Marche (UNIVPM), Italy \\
    \textsuperscript{\rm 3}Weizmann Institute of Science, Rehovot, Israel \quad
    \textsuperscript{\rm 4}Bocconi University, Italy \\
    \{aprocopio, andreygizdov\}@fas.harvard.edu, \;
    \{m.esposito, s.raggiunto, a.belli, p.pierleoni\}@staff.univpm.it
}

\begin{document}

\maketitle

\begin{abstract}
We study on-device time-series analysis for gait detection in Parkinson’s disease (PD) from short windows of triaxial acceleration, targeting resource-constrained wearables and edge nodes. We compare magnitude thresholding to three 1D CNNs for time-series analysis: a literature baseline (separable convolutions) and two ultra-light models—one purely separable and one with residual connections. Using the ``BioStampRC21'' dataset, 2~s windows at 30~Hz, and subject-independent leave-one-subject-out (LOSO) validation on 16 PwPD with chest-worn IMUs, our residual separable model (\emph{Model 2}, 533 params) attains PR--AUC = 94.5\%, $F_1 = 91.2\%$, MCC = 89.4\%, matching or surpassing the baseline (5{,}552 params; PR--AUC = 93.7\%, $F_1 = 90.5\%$, MCC = 88.5\%) with ${\sim}10{\times}$ fewer parameters. The smallest model (\emph{Model 1}, 305 params) reaches PR--AUC = 94.0\%, $F_1 = 91.0\%$, MCC = 89.1\%. Thresholding obtains high recall (89.0\%) but low precision (76.5\%), yielding many false positives and high inter-subject variance. Sensor-position analysis (train-on-all) shows chest and thighs are most reliable; forearms degrade precision/recall due to non-gait arm motion; naive fusion of all sites does not outperform the best single site. Both compact CNNs execute within tight memory/latency budgets on STM32-class MCUs (sub-10~ms on low-power boards), enabling on-sensor gating of transmission/storage. Overall, ultra-light separable CNNs provide a superior accuracy--efficiency--generalization trade-off to fixed thresholds for wearable PD gait detection and underscore the value of tailored time-series models for edge deployment.
\end{abstract}

\section{Introduction}
Parkinson's disease (PD) is now one of the fastest-growing neurodegenerative disorders in the world 
\cite{chaudhuri2024economic}. Despite optimal treatment, the progression of the disease leads to a gradual deterioration in motor and cognitive disorders, with inevitable compromise in patients’ quality of life. 
Several works in the literature have studied gait disabilities using wearable, small, and non-invasive sensors. To design personalized interventions and overcome the limitations related to self-reported questionnaires and diaries, the focus is to improve clinical and home care through the additional information on daily life performance provided by Inertial Measurement Units (IMUs) \cite{corra2021comparison}. 
It is essential to develop algorithms for rapid and precise detection of motor activity in subjects with Parkinson's disease to intervene promptly in the detection of anomalies to safeguard the subject's health.  To carry out continuous and constant monitoring of PD subjects, particularly in a domestic and unsupervised environment, the first fundamental step is the detection of the activities carried out by the subject \cite{muthukrishnan2020wearable, corra2021comparison}. In particular, gait activities are of interest due to the fact that PD severely affects the gait cycle and, therefore, gait cycle impairments should be monitored and detected promptly. Following gait detection, it is advisable to perform gait analysis to detect all possible anomalies such as fluctuations, dyskinesias, etc. \cite{sigcha2020deep, mancini2021measuring}. The main studies either process the signal offline or use Cloud-based services. Cloud-based architectures need data to be moved to data centers, and are typically employed to support long-term data analyses. However, this scenario is contrary to IoT application requirements, such as limiting costs, memory footprint, processing, and communication resources \cite{kianoush2023random}. In recent years, there has been a shift towards edge computing, which means processing data on devices closer to the data sources, rather than relying on cloud-based services to reduce the amount of data to be sent to the cloud by the IoT devices \cite{fathalla2022lstm}. This shift has also been enabled by advances in embedded  systems, as well as tools and accelerators that make it possible to run machine learning and artificial intelligence algorithms on low-power devices \cite{dini2024overview}.\\

\noindent In this paper, an in-depth investigation of gait recognition techniques is conducted as a first step towards developing a decentralized architecture for Parkinson’s disease monitoring where distributed processing happens across wearable devices, edge nodes, and the cloud. The proposed gait recognition algorithm aims to reduce data transmission through local data processing, optimizing the system’s energy efficiency and improving the quality of stored data by eliminating redundant or irrelevant data samples. Compared to previous works, we provide a comprehensive evaluation that compares simple threshold-based methods to separable CNNs-based frameworks, including a Residual Network structure. Our evaluation covers multiple sensor positions and combinations, assesses the methods' performance across different patients (Leave-One-Subject-Out validation), and examines computational complexity and inference time on a target MCU.\\

In this work, we make the following contributions:

\begin{enumerate}
    \item   
          We introduce two lightweight configurations of 1D-CNNs-based architectures achieving a better complexity-performance tradeoff and generalization via F1-score compared to a reference, larger baseline. Through the LOSO evaluation, we show that it is possible to design a super-light architecture that matches the performance of models 18× larger while having extremely light inference times in tasks of binary gait detection on microcontrollers.
          
    \item 
          We discuss flaws of a threshold-based approach, mainly concerning an inability for the framework to generalize among different subjects and sensor positions, and to maintain a stable accuracy level.

    \item 
          We conduct an in-depth investigation of models' performances and generalization abilities based on the position of data acquisition sensors. We show that data acquired from accelerometers placed in the chest area is consistently less noisy and easier to classify for all the frameworks evaluated, while thighs, and in particular forearms, appear to show greater inconsistency and variance.
    
\end{enumerate}

\section{Methods}
\label{sec:methods}

Our goal is to investigate optimal architectures for gait recognition via efficient wearable sensors. Gait recognition can be utilized for contextual activation of efficient data transmission/storage to limit bandwidth and power consumption, which is especially useful for daily monitoring. 

\subsection{Data and Preprocessing}
\subsubsection{Dataset description}
The raw data utilized in this study was obtained from the ``BioStampRC21'' dataset (hereafter, BioStamp) by \cite{PD-BIOSTAMP-RC21}, which is accompanied by clinical metadata including patient IDs and health statuses (patients with Parkinson’s disease (PwPD) or Control). BioStamp gathers data from 17 PwPD and 17 control subjects of matching ages over an average time of 45.4 hours. The PwPD sample consists of 7 women and 10 men with a mean age of 66.4 years (11.3 SD), while the age-matched sample consists of 13 women and 4 men with a mean age of 64.0 years (9.9 SD). Acceleration data in each individual subject is recorded from 5 different sensor positions: chest (ch), left thigh (ll), right thigh (rl), left forearm (lh), right forearm (rh). A clinical annotation file for each subject provides a description of the clinical test being conducted at each timestamp during data collection.

\subsubsection{Preprocessing}
Data annotations were categorized into ``gait'' and ``non-gait'' events based on the nature of the clinical test. For example, gait-related activities included UPDRS and UHDRS test versions of ``Timed Up and Go'', ``10 Meter Walk Test'', ``Gait'', and ``Tandem Walking''. Events that did not comprise the selected gait activities were categorized as non-gait. The only annotation excluded from both criteria was the ``UHDRS 15 - Retropulsion Pull Test'', due to ambiguity in classifying the event and a negligible relevance both in the dataset and for the representation of domestic activities in PwPD. Sensor data were sourced from accelerometers placed at the five previously described body positions: ch, ll, rl, lh, rh. Each participant's sensor data underwent column standardization, renaming original columns to maintain consistency across the dataset. Subsequently, the accelerometer readings were uniformly resampled from 31.25 Hz to 30 Hz to allow the use of integer-length windows.

\subsubsection{Windowing}
Data corresponding to gait and non-gait intervals were segmented using timestamps derived from clinical annotations. Each sensor recording was filtered according to these intervals to isolate relevant data segments. Windows of accelerometer data were created using a sliding-window approach with a window size of 60 samples (2 seconds) and a step size of 15 samples, resulting in 75\% overlap uniformly across both classes. Each windowed data segment was standardized by subtracting the mean, ensuring zero-centered data for analysis. Finally, data segments were labeled according to the sensor position, the type of activity (gait or non-gait) and the participant's clinical status (PD or Control).

\subsubsection{Final datasets}
Processed data from each sensor were compiled into structured dictionaries, organized by participant IDs and sensor positions. This led to the creation of 5 datasets, based on sensor positions. Subject ``ID38'' was excluded during the dataset generation process due to data quality issues. Acceleration data from all 5 sensors were aggregated across each participant and concatenated to form a comprehensive dataset. Each dataset (the 5 per-sensor datasets and the comprehensive dataset) was divided into training set (60\%), a validation set (20\%), and test set (20\%) using stratified random splitting to maintain a consistent label distribution. The stratification ensured balanced representation of gait and non-gait labels in both training and validation datasets.

\subsubsection{Leave-One-Subject-Out (LOSO)}
To prepare the dataset for LOSO cross-validation, only PwPD were selected for the analysis, resulting in a final pool of 16 subjects: IDs = [6, 10, 12, 13, 15, 17, 23, 24, 25, 33, 35, 36, 40, 42, 44, 63]. For each of the resulting 16 PwPD, data from that participant served as the test set, while the remaining participants' data constituted training (70\%) and validation (30\%) datasets. Accelerometer data only from the chest sensor were extracted for this analysis as they proved to be the most stable from our sensor-dependent analysis. Every dataset obtained – LOSO, sensor-specific and combined – is composed of windows of shape $60 \times 3$, representing the 2-second windows with triaxial accelerometer data sampled at 30 Hz.

\subsection{Gait detection algorithms}
\label{sec:Gait detection algorithms}
\subsubsection{Magnitude thresholding}
While AI methods offer replicability and generalization, simple thresholding has proven able to distinguish between gait and non-gait while still maintaining extremely low computational cost \citep{Borzì-2023, soumma2024self}. As such, a comparison between the 1D-CNN models and a threshold-based detector is carried out. Thresholding is applied to the magnitude of the acceleration signal. 
For the $j^{th}$ window, magnitude is computed as follows:
\[
A_j = \sum_{i=1}^{n} \left( a_{x_i}^2 + a_{y_i}^2 + a_{z_i}^2 \right)
\]
where $n$ is the number of samples in the window, and $a_{x_i}$, $a_{y_i}$, and $a_{z_i}$ are the measured acceleration components along the 3 measurement axes for the $i^{th}$ time step. The identification of gait windows is performed by evaluating whether the magnitude of the window is above the threshold. Consistent with previous works \citet{Borzì-2023}, the optimal threshold is selected as the value that maximizes the F1 score across the training dataset. In the evaluation process, all windows with magnitude higher than the optimal threshold are classified as gait, and conversely the windows with a lower magnitude are classified as non-gait.
\subsection{Deep Learning frameworks with CNNs.}
In this section we describe and evaluate three different one-dimensional Convolutional Neural Network (1D-CNNs) architectures for time-series analysis. 

\paragraph{Reference 1D-CNN Baseline}
Following \citet{Borzì-2023}, who adapt the DeepFog CNN of \citet{bikias2021deepfog} by replacing standard 1D convolutions with separable 1D convolutions to reduce complexity, we adopt the same two-block 1D SepConv CNN as our baseline, instantiated at our window length (60 samples) and with a two-class softmax head. Table \ref{tab:baseline_model} shows the architecture. Both SepConv layers include bias terms and use acausal padding before the ReLU activation.

\noindent\textit{Parameter profile:}
SepConv1: $430$ params; SepConv2: $5{,}040$; Dense: $82$; total ${5{,}552}$.

\begin{table}[!t]
\centering
\begin{tabular}{@{}ll@{}}
\toprule
\textbf{Component} & \textbf{Configuration} \\
\midrule
Input & $(60,3)$ \\
SepConv1 & filters: $100$, kernel: $10$, activation: ReLU \\
Pooling & MaxPool1D (pool size: $3$): length $60 \rightarrow 20$ \\
SepConv2 & filters: $40$, kernel: $10$, activation: ReLU \\
Aggregation & GlobalAvgPooling1D, output dim: $40$ \\
Regularization & Dropout, $p=0.5$ \\
Classifier & Dense ($2$), softmax \\
\bottomrule
\end{tabular}
\caption{Reference baseline architecture.}
\label{tab:baseline_model}
\end{table}

\paragraph{Model 1: Super-Light Separable Convolutions Network}
The first architecture we propose is a highly compact variant of the baseline (retaining depthwise and pointwise separable 1D convolutions) but with a very reduced parameter count: fewer filters (8 and 16 instead of 100 and 40), smaller kernels (5 and 7 instead of 10), less aggressive pooling ($\times 2$ instead of $\times 3$), Batch Normalization (BN) after each convolution with biases disabled, and the same global average pooling + softmax head.  Furthermore, we adopt an increasing number of channels per layer (8$\rightarrow$16) rather than the baseline’s decreasing one (100$\rightarrow$40). This follows the common CNN practice of widening the network with depth to capture more complex features and mitigate information bottlenecks in later layers \citep{SimonyanZisserman2014,HeZhangRenSun2016,Sandler2018}.

\noindent\textit{Parameter profile}
Params (including BN $\gamma,\beta$): SepConv1: $39$; BN1: $16$; SepConv2: $184$; BN2: $32$; Dense: $34$; total $\mathbf{305}$.\\

\begin{table}[!t]
\centering
\begin{tabular}{@{}ll@{}}
\toprule
\textbf{Component} & \textbf{Configuration} \\
\midrule
Input & $(60,3)$ \\
SepConv1 & filters: $8$, kernel: $5$ \\
Post-processing & BN $\rightarrow$ ReLU $\rightarrow$ MaxPool1D (pool size: $2$)\\
SepConv2 & filters: $16$, kernel: $7$ \\
Post-processing & BN $\rightarrow$ ReLU \\
Aggregation & GlobalAvgPooling1D, output dim: $16$ \\
Classifier & Dense ($2$), softmax \\
Initialization & He normal (depthwise and pointwise) \\
\bottomrule
\end{tabular}
\caption{Model~1 architecture.}
\label{tab:model1_hparams}
\end{table}

\paragraph{Model 2: Super-Light Residual Separable CNN}
The second architecture we analyze, Model~2, keeps the same lightweight design footprint as Model~1 (separable 1D convolutions and two stages) but changes two aspects: (i) each stage is a \emph{residual} block with an identity skip (projected with a $1{\times}1$ convolution when channel dimensions differ), and (ii) both convolutions use a larger kernel ($k{=}9$). The first block downsamples with \emph{average} pooling (factor $2$); the second does not downsample. Each block has convolutions with bias disabled, followed by BatchNorm and ReLU; the residual sum is followed by a ReLU. A global average pooling layer and a two-class softmax head complete the model.

\noindent\textit{Parameter profile}
With input $(60,3)$ the trainable parameters are: Block~1 (SepConv+BN+projection) $=107$, Block~2 $=392$, Dense head $=34$, totaling $\mathbf{533}$ parameters.

\begin{table}[!t]
\centering
\begin{tabular}{@{}ll@{}}
\toprule
\textbf{Component} & \textbf{Configuration} \\
\midrule
Input & $(60,3)$ \\
ResBlock1 (main) & SepConv1D: filters: $8$, kernel: $9$ \\
Post-processing & BN $\rightarrow$ ReLU $\rightarrow$ AvgPool1D (size: $2$) \\
ResBlock1 (skip) & $1{\times}1$ Conv $+$ BN ($3 \rightarrow 8$) \\
Residual & Add $\rightarrow$ ReLU \\
ResBlock2 (main) & SepConv1D: filters: $16$, kernel: $9$ \\
Post-processing & BN $\rightarrow$ ReLU \\
ResBlock2 (skip) & $1{\times}1$ Conv $+$ BN ($8 \rightarrow 16$) \\
Residual & Add $\rightarrow$ ReLU \\
Aggregation & GlobalAvgPooling1D, output dim: $16$ \\
Classifier & Dense ($2$), softmax \\
Initialization & He normal (depthwise, pointwise, $1{\times}1$) \\
\bottomrule
\end{tabular}
\caption{Model~2 architecture.}
\label{tab:model2_hparams}
\end{table}

\subsection{Training Protocol}
All models share the same pipeline and training hyperparameters. Due to compatibility requirements all models have Softmax heads and use categorical cross-entropy and one-hot labels.

\paragraph{Data split (LOSO)}
We use Leave-One-Subject-Out (LOSO) evaluation. For each held-out subject $s$, we train on the remaining subjects, split a validation subset from the training portion (as provided by the dataset loader), and test on $s$. All choices below are applied identically in every fold and to every model.

\paragraph{Optimization}
We train using AdamW, \citet{LoshchilovHutter2019} – Adaptive Moment Estimator with Decoupled Weight Decay – learning rate $1{\times}10^{-3}$, weight decay $1{\times}10^{-4}$, $\beta_1{=}0.9$, $\beta_2{=}0.99$, $\epsilon{=}10^{-8}$. Batch normalization and bias parameters are excluded from weight decay. While training is capped at 300 epochs with batch size 256, we monitor validation PR-AUC (Average Precision) and use Early Stopping with patience 12, restoring the best weights. We apply Reduce Learning Rate on Plateau on the same metric (factor $0.2$, patience 7, minimum learning rate $1{\times}10^{-5}$).
Furthermore, we apply class-balanced weights based on the \emph{effective number of samples} \citep{Cui2019ClassBalanced}. Let $n_c$ be the number of training samples for class $c \in \{0,1\}$ and $\beta{=}0.999$. The effective number is $E(n_c) = \frac{1-\beta^{n_c}}{1-\beta}$; the unnormalized weight is $w_c' = \frac{1}{E(n_c)} = \frac{1-\beta}{1-\beta^{n_c}}$. We normalize to unit average weight,
\[w_c = \frac{w_c'}{\;\frac{w_0' n_0 + w_1' n_1}{n_0 + n_1}\;},\]
and pass $(w_0,w_1)$ during training for loss computation.\\
After training in each fold, we compute positive-class scores on the validation set and select a single threshold $\tau^{*}$ that maximizes validation F1 by sweeping 401 uniformly spaced values in $[0,1]$. We then \emph{fix} $\tau^{*}$ and evaluate the test subject with no further tuning.
\textit{Note on reproducibility:}
Implementation uses TensorFlow--Keras (v2.15) and scikit-learn metrics for PR-AUC and F1. A fixed random seed (n=12) is set for all the training scripts. All training and testing settings are identical across models. 
\paragraph{Deployment evaluation} After training, they were exported on the ST Edge AI Developer Cloud tool, which provides a framework for the evaluation of DL models for their deployment on various STM32 boards and MCUs. In particular, it provides estimations on memory occupancy and inference time for a given target MCU.

\subsection{Evaluation Metrics}
The dataset is imbalanced (about 80\% non-gait, 20\% gait). We therefore prioritize metrics that are informative under class imbalance and report both threshold-free and thresholded performance. All metrics are computed per LOSO fold; tables report the mean across folds (per-subject values in the supplement).

\paragraph{Notation}
Let $\mathrm{TP}, \mathrm{TN}, \mathrm{FP}, \mathrm{FN}$ denote the confusion counts for the positive class (gait), then we compute the following.
PR-AUC (Average Precision), i.e. the area under the precision–recall curve (computed as Average Precision); threshold-free and robust to class imbalance.
F1 at fixed threshold, i.e., the harmonic mean of precision and recall at the model’s fixed decision threshold $\tau$, selected on validation by maximizing F1:
\[\mathrm{F1} \;=\; \frac{2\,\mathrm{PPV}\,\mathrm{TPR}}{\mathrm{PPV}+\mathrm{TPR}},
\quad \mathrm{PPV} \;=\; \frac{\mathrm{TP}}{\mathrm{TP}+\mathrm{FP}}\]
\[\mathrm{TPR} \;=\; \frac{\mathrm{TP}}{\mathrm{TP}+\mathrm{FN}},
\quad \mathrm{TNR} \;=\; \frac{\mathrm{TN}}{\mathrm{TN}+\mathrm{FP}}.\]
Furthermore, in the summary table \ref{tab:LOSO_avg} we include the Geometric Mean of TPR and TNR (GM), the chance-corrected agreement (Cohen's K), the Accuracy (kept for reference but de-emphasized under class imbalance), and the Matthews correlation coefficient (MCC), which provides a single summary that captures whether a model balances both classes rather than succeeding by majority-class guessing.

\[ \mathrm{Accuracy} \;=\; \frac{\mathrm{TP}+\mathrm{TN}}{\mathrm{TP}+\mathrm{TN}+\mathrm{FP}+\mathrm{FN}} \]\\
\[ \mathrm{MCC} \;=\; \frac{\mathrm{TP}\,\cdot\mathrm{TN}-\mathrm{FP}\,\cdot\mathrm{FN}} {\sqrt{(\mathrm{TP}+\mathrm{FP})(\mathrm{TP}+\mathrm{FN})(\mathrm{TN}+\mathrm{FP})(\mathrm{TN}+\mathrm{FN})}} \]\\

\paragraph{Reporting}
For each fold we fix the threshold $\tau^{*}$ on validation by maximizing F1, then evaluate the test subject with that $\tau^{*}$. We report PR-AUC (threshold-free), and at $\tau^{*}$ we report F1, Precision (PPV), Recall (TPR), Specificity (TNR), MCC, and Accuracy.\\

\section{Results}
\label{sec:results}
\
\subsection{Subject-Independent LOSO on Chest IMU}
Table~\ref{tab:LOSO_avg} reports LOSO means over 16 PwPD using the chest IMU (2\,s, 60 samples, 3 axes). Both proposed compact CNNs outperform the Baseline model while using $90$--$95\%$ fewer parameters: \emph{Model~2} reaches F1=91.2\%, PR--AUC=94.5\%, MCC=89.4\%; \emph{Model~1} attains F1=91.0\%, PR--AUC=94.0\%, MCC=89.1\%. The Baseline architecture yields F1=90.5\%, PR--AUC=93.7\%, MCC=88.5\%. Gains vs.\ Baseline are small but consistent across PPV (Model~2: $94.0$ vs.\ $93.3$) and TPR (Model~2: $89.1$ vs.\ $88.3$).
MCC trends mirror F1/PR--AUC, indicating a better balance of false positives/negatives under the $\approx$80/20 non-gait/gait skew.\\

\begin{table}[!htbp]
\centering
\caption{Average LOSO performance.}
\label{tab:LOSO_avg}
\footnotesize
\begin{tabular}{lrrrr}
\toprule
Metric \% & Model 1 & Model 2 & Baseline & Threshold \\
\midrule
Accuracy & 96.6 & 96.7 & 96.4 & 92.4\\
Specificity & 98.5 & 98.6 & 98.5 & 93.6\\
Recall & 88.9 & 89.1 & 88.3 & 89.0\\
Precision & 93.7 & 94.0 & 93.3 & 76.5\\
F1-score & 91.0 & 91.2 & 90.5 & 81.5\\
MCC & 89.1 & 89.4 & 88.5 & 77.7\\
GM & 93.5 & 93.6 & 93.1 & 90.9\\
Cohen-K & 88.9 & 89.2 & 88.3 & 76.8\\
PR-AUC & 94.0 & 94.5 & 93.7 & --- \\
\bottomrule
\end{tabular}
\end{table}

\noindent The threshold baseline has good recall but low precision and is unstable across subjects. Averaged over LOSO (Table~\ref{tab:LOSO_avg}), it achieves TPR=\(89.0\%\) but only PPV=\(76.5\%\), for F1=\(81.5\%\) and MCC=\(77.7\%\). Per-subject variability is high (F1 $\sigma=9.9$; Table~\ref{tab:LOSO_threshold}), reflecting numerous false positives on some patients.
Generalization across positions is also weak in the subject-mixed setting (Table~\ref{tab:sensor_specific_evaluation}): on forearms, PPV falls behind despite high recall (Left Arm: PPV=43.0\%, F1=57.1\%; Right Arm: PPV=42.2\%, F1=57.0\%), possibly due to non-gait arm motions that can mimic the periodicity of motion.
Despite having only 305 (Model~1) and 533 (Model~2) parameters compared to the Baseline's 5,552, both compact models match or surpass it on LOSO chest. Relative to Baseline, Model~2 gains \(+0.7\) F1, \(+0.8\) PR--AUC, and \(+0.9\) MCC; Model~1 gains \(+0.5\) F1, \(+0.3\) PR--AUC, \(+0.6\) MCC. Versus the threshold method, Model~2 improves PPV by \(+\ 17.5\) points (94.0 vs.\ 76.5) at similar recall (89.1 vs.\ 89.0), reducing false positives without relevant effects in sensitivity. 
Tables~\ref{tab:LOSO_model1}--\ref{tab:LOSO_baseline} detail per-subject results, also showing standard deviation ($\sigma$) and average values ($\mu$) for all metrics. Difficult cases appear for all CNNs (e.g., IDs 25, 33, 6), where TPR is lowered while PPV remains high. Several subjects approach maximum values across all metrics (e.g., IDs 36, 40, 15, 13). Thresholding, although in general worse across metrics, achieves better results for the same subjects. However, it exhibits larger variance across metrics (F1 $\sigma=9.9$)) with very unstable PPV, showing a higher susceptibility to false positives. While the simplest and most lightweight method, it is therefore not suitable for applications where misclassification cost is high (e.g., when it is needed to transmit or store gait windows).

\noindent Table~\ref{tab:sensor_specific_evaluation}
reports per-sensor performance, with all models trained on the full dataset and evaluated on each position. Chest and thighs are strongest across models. For the residual separable CNN (Model~2), left-thigh is best (F1 = 88.7\%), followed by chest (F1 = 87.5\%) and right-thigh (F1 = 84.7\%). The lightweight separable CNN (Model~1) follows the same pattern (left-thigh 88.5\%, chest 84.2\%). The literature baseline also peaks on the left-thigh (89.1\%) and chest (88.4\%). Forearms underperform for all CNNs (Model~2: 76.0--81.8\% F1), and the threshold baseline shows the highest precision drop on forearms (PPV = 43.0\%/42.2\%; F1 = 57.1\%/57.0\% for left and right) despite high recall. The Threshold method performs markedly worse when tested on the fused 'All' sensor data (F1 = 67.7\%) compared to the CNN models (Model 2 F1 = 83.8\%), indicating its poor generalization across sensors.

\begin{table}[!t]
\centering
\caption{LOSO evaluation for Model 1.}
\label{tab:LOSO_model1}
\footnotesize
\begin{tabular}{lrrrrr}
\toprule
ID & Acc (\%) & TNR (\%) & TPR (\%) & PPV (\%) & F1 (\%) \\
\midrule
        6 & 92.8 & 97.8 & 78.2 & 92.7 & 84.8 \\ 
        10 & 98.0 & 100.0 & 90.8 & 100.0 & 95.2 \\ 
        12 & 95.0 & 96.8 & 86.6 & 84.6 & 85.6 \\ 
        13 & 98.5 & 99.8 & 92.4 & 98.8 & 95.5 \\ 
        15 & 96.8 & 98.3 & 91.8 & 94.2 & 93.0 \\ 
        17 & 97.9 & 98.9 & 93.4 & 95.1 & 94.2 \\ 
        23 & 98.3 & 99.6 & 91.7 & 97.8 & 94.7 \\ 
        24 & 98.2 & 99.2 & 89.3 & 92.6 & 90.9 \\ 
        25 & 93.4 & 97.6 & 78.1 & 90.2 & 83.7 \\
        33 & 95.2 & 99.6 & 77.2 & 97.9 & 86.4 \\ 
        35 & 99.1 & 99.7 & 95.8 & 98.3 & 97.0 \\ 
        36 & 99.9 & 100 & 99.0 & 100.0 & 99.5 \\ 
        40 & 96.1 & 98.4 & 84.1 & 91.2 & 87.5 \\
        42 & 95.2 & 98.7 & 84.1 & 95.4 & 89.4 \\ 
        44 & 94.4 & 94.5 & 93.8 & 79.1 & 85.8 \\ 
        63 & 97.2 & 97.6 & 95.4 & 91.3 & 93.3 \\ 
\midrule
$\mu$    & 96.6 & 98.5 & 88.9 & 93.7 & 91.0 \\
$\sigma$ &  2.1 &  1.4 &  6.8 &  5.7 &  4.9 \\
\bottomrule
\end{tabular}
\end{table}

\begin{table}[!t]
\centering
\caption{LOSO evaluation for Model 2 (ResNet).}
\label{tab:LOSO_model2}
\footnotesize
\begin{tabular}{lrrrrr}
\toprule
ID & Acc (\%) & TNR (\%) & TPR (\%) & PPV (\%) & F1 (\%) \\
\midrule
        6 & 92.1 & 98.0 & 75.1 & 93.1 & 83.2 \\ 
        10 & 97.9 & 99.9 & 90.8 & 99.6 & 95.0 \\ 
        12 & 96.2 & 98.1 & 86.9 & 90.5 & 88.6 \\ 
        13 & 98.7 & 99.5 & 94.7 & 97.6 & 96.1 \\ 
        15 & 98.1 & 98.9 & 95.7 & 96.2 & 95.9 \\ 
        17 & 98.1 & 99.1 & 94.1 & 95.9 & 95.0 \\ 
        23 & 97.8 & 99.3 & 90.3 & 96.3 & 93.2 \\
        24 & 98.9 & 99.3 & 95.2 & 94.1 & 94.7 \\ 
        25 & 91.3 & 97.9 & 67.6 & 90.2 & 77.3 \\ 
        33 & 95.7 & 99.7 & 79.3 & 98.5 & 87.8 \\ 
        35 & 99.2 & 99.8 & 95.8 & 99.1 & 97.4 \\ 
        36 & 99.6 & 99.5 & 100.0 & 97.1 & 98.5 \\
        40 & 96.9 & 98.3 & 89.8 & 90.9 & 90.4 \\ 
        42 & 95.8 & 99.3 & 84.6 & 97.5 & 90.6 \\ 
        44 & 93.5 & 94.3 & 89.9 & 77.6 & 83.3 \\
        63 & 96.7 & 97.1 & 95.2 & 89.5 & 92.3 \\
\midrule
$\mu$    & 96.7 & 98.6 & 89.1 & 94.0 & 91.2 \\
$\sigma$ &  2.5 &  1.4 &  8.6 &  5.5 &  5.9 \\
\bottomrule
\end{tabular}
\end{table}

\begin{table}[!t]
\centering
\caption{LOSO evaluation for the Baseline model.}
\label{tab:LOSO_baseline}
\footnotesize
\begin{tabular}{lrrrrr}
\toprule
ID & Acc (\%) & TNR (\%) & TPR (\%) & PPV (\%) & F1 (\%) \\
\midrule
        10 & 91.2 & 98.0 & 71.8 & 92.8 & 80.9 \\ 
        12 & 97.8 & 100.0 & 89.7 & 100.0 & 94.6 \\ 
        13 & 94.4 & 96.6 & 83.7 & 83.5 & 83.6 \\ 
        15 & 98.7 & 99.8 & 92.8 & 99.2 & 95.9 \\ 
        17 & 97.0 & 97.4 & 95.4 & 91.6 & 93.5 \\ 
        23 & 98.3 & 99.1 & 95.2 & 95.9 & 95.6 \\
        24 & 97.9 & 99.4 & 90.3 & 97.0 & 93.6 \\ 
        25 & 98.3 & 98.9 & 92.9 & 90.7 & 91.8 \\ 
        33 & 92.5 & 97.1 & 76.1 & 87.9 & 81.6 \\ 
        35 & 95.4 & 99.3 & 79.3 & 96.5 & 87.1 \\ 
        36 & 99.4 & 99.8 & 96.6 & 99.1 & 97.9 \\ 
        40 & 99.9 & 100 & 99.0 & 100 & 99.5 \\ 
        42 & 96.1 & 98.8 & 82.1 & 93.1 & 87.3 \\ 
        44 & 95.0 & 99.0 & 82.7 & 96.2 & 89.0 \\ 
        6 & 94.4 & 95.4 & 89.9 & 81.1 & 85.3 \\ 
        63 & 96.3 & 96.5 & 95.2 & 87.7 & 91.3 \\ 
\midrule
$\mu$    & 96.4 & 98.5 & 88.3 & 93.3 & 90.5 \\
$\sigma$ &  2.5 &  1.4 &  8.0 &  5.9 &  5.8 \\
\bottomrule
\end{tabular}
\end{table}

\begin{table}[!t]
\centering
\caption{LOSO evaluation for the Threshold baseline.}
\label{tab:LOSO_threshold}
\footnotesize
\begin{tabular}{lrrrrr}
\toprule
ID & Acc (\%) & TNR (\%) & TPR (\%) & PPV (\%) & F1 (\%) \\
\midrule
6 & 83.4 & 93.4 & 55.0 & 74.3 & 63.2 \\ 
10 & 94.2 & 95.2 & 90.8 & 83.8 & 87.2 \\ 
12 & 91.2 & 91.5 & 89.7 & 68.3 & 77.5 \\ 
13 & 95.2 & 95.0 & 96.2 & 79.3 & 86.9 \\ 
15 & 96.7 & 95.8 & 99.7 & 87.6 & 93.3 \\ 
17 & 93.9 & 92.8 & 98.9 & 76.0 & 85.9 \\ 
23 & 96.4 & 98.2 & 87.6 & 90.7 & 89.1 \\ 
24 & 97.5 & 97.2 & 100 & 80.0 & 88.9 \\ 
25 & 84.5 & 93.5 & 52.0 & 69.1 & 59.3 \\
33 & 90.3 & 92.5 & 81.3 & 72.5 & 76.6 \\ 
35 & 92.9 & 92.2 & 96.6 & 69.1 & 80.6 \\
36 & 96.6 & 96.0 & 100 & 81.1 & 89.6 \\ 
40 & 94.1 & 94.3 & 93.1 & 76.1 & 83.7 \\
42 & 92.9 & 93.9 & 89.7 & 82.4 & 85.9 \\ 
44 & 84.6 & 81.9 & 96.9 & 54.1 & 69.4 \\ 
63 & 94.0 & 93.6 & 95.7 & 79.4 & 86.8 \\ 
\midrule
$\mu$ & 92.4 & 93.6 & 89.0 & 76.5 & 81.5 \\
$\sigma$ & 4.5 & 3.6 & 14.8 & 8.8 & 9.9 \\
\bottomrule
\end{tabular}
\end{table}

\subsection{Compute efficiency and deployability}
We profile three STM32 targets (L4R9I-DISCO, F401RE, H7S78-DK). Table~\ref{tab:consolidated_metrics} reports memory, MACs, and per-window latency. The Baseline is largest and slowest (133{,}755 MACs; 13.75\,KB RAM; H7: 3.22\,ms; L4/F4: 26.68/32.72\,ms). Model~1 is the most compact (9{,}879 MACs; 4.87\,KB RAM) and fastest (H7: 0.45\,ms; L4/F4: 3.42/4.18\,ms). Model~2 offers the best accuracy–efficiency balance (18{,}303 MACs; H7: 0.70\,ms; L4/F4: 5.50/6.80\,ms), delivering slightly higher $F_1$ than Baseline with $\sim\!7.3\times$ fewer MACs. All models fit without external flash/RAM, enabling on-sensor gating. Because dynamic energy on MCUs scales roughly with operation count, lower MACs imply lower energy per inference \cite{li2025qpart}. By contrast, fixed-threshold detection has a cost that grows linearly with input length and is effectively negligible on our windows \cite{Borzì-2023}, but its poor precision and high inter-subject variance (Table~\ref{tab:LOSO_threshold}) can make downstream false-positive costs (radio, storage, block activation) dominate.

\begin{table}[!t]
\centering
\caption{Model Metrics and Inference Latency}
\label{tab:consolidated_metrics}
\renewcommand{\arraystretch}{1.2}
\begin{adjustbox}{max size={\columnwidth}{\textheight}, center}
\begin{tabular}{ccccccr}
\toprule
\multirow{2}{*}{\textbf{Model}} & \textbf{Flash} & \textbf{RAM} & \textbf{MACs} & \multicolumn{3}{c}{\textbf{Inference Latency (ms)}} \\
\cline{5-7}
 & \textbf{(KB)} & \textbf{(KB)} & & \textbf{L4R9I-DISCO} & \textbf{F401RE} & \textbf{H7S78-DK} \\
\midrule
\midrule
Model 1 & 20.62 & 4.87 & 9879  & 3.42 & 4.18 & 0.45 \\
Model 2 & 23.86 & 11.93 & 18303 & 5.50 & 6.80 & 0.70 \\
Baseline & 42.12 & 13.75 & 133755 & 26.68 & 32.72 & 3.22 \\
\bottomrule
\end{tabular}
\end{adjustbox}
\end{table}

\begin{table}[!t]
\centering
\caption{\textbf{Train-on-all, per-sensor evaluation (subject-mixed).} TNR = specificity; TPR = recall; PPV = precision.}
\label{tab:sensor_specific_evaluation}
\footnotesize
\begin{tabular}{@{}llrrrrr@{}}
\toprule
\textbf{Model} & \textbf{Sensor} & \textbf{Acc} & \textbf{TNR} & \textbf{TPR} & \textbf{PPV} & \textbf{F1} \\
\midrule
\multirow{6}{*}{Model 1}
  & Chest      & 94.5 & 99.5 & 74.2 & 97.3 & 84.2 \\
  & Left arm   & 91.4 & 96.1 & 72.0 & 81.8 & 76.6 \\
  & Right arm  & 92.4 & 96.3 & 76.7 & 83.5 & 80.0 \\
  & Left leg   & 95.8 & 99.0 & 82.6 & 95.3 & 88.5 \\
  & Right leg  & 94.1 & 98.6 & 75.6 & 93.2 & 83.4 \\
  & All        & 93.7 & 97.9 & 76.3 & 90.0 & 82.5 \\
\midrule
\multirow{6}{*}{Model 2}
  & Chest      & 95.5 & 99.1 & 80.5 & 95.8 & 87.5 \\
  & Left arm   & 91.3 & 96.3 & 70.7 & 82.3 & 76.0 \\
  & Right arm  & 93.2 & 97.1 & 77.3 & 86.9 & 81.8 \\
  & Left leg   & 95.8 & 98.7 & 83.8 & 94.2 & 88.7 \\
  & Right leg  & 94.5 & 98.6 & 77.6 & 93.1 & 84.7 \\
  & All        & 94.1 & 98.0 & 78.1 & 90.5 & 83.8 \\
\midrule
\multirow{6}{*}{Baseline}
  & Chest      & 95.8 & 99.1 & 81.9 & 95.9 & 88.4 \\
  & Left arm   & 92.0 & 96.9 & 71.9 & 84.8 & 77.8 \\
  & Right arm  & 93.5 & 97.8 & 76.2 & 89.4 & 82.3 \\
  & Left leg   & 96.0 & 98.9 & 83.8 & 95.1 & 89.1 \\
  & Right leg  & 94.9 & 98.8 & 78.7 & 94.2 & 85.7 \\
  & All        & 94.4 & 98.3 & 78.6 & 92.0 & 84.8 \\
\midrule
\multirow{6}{*}{Threshold}
  & Chest      & 91.7 & 95.0 & 78.3 & 79.2 & 78.8 \\
  & Left arm   & 75.1 & 72.7 & 84.9 & 43.0 & 57.1 \\
  & Right arm  & 73.9 & 70.6 & 87.6 & 42.2 & 57.0 \\
  & Left leg   & 90.1 & 90.0 & 90.7 & 68.9 & 78.3 \\
  & Right leg  & 88.2 & 88.3 & 87.9 & 64.8 & 74.6 \\
  & All        & 83.9 & 83.4 & 85.9 & 55.8 & 67.7 \\
\bottomrule
\end{tabular}
\end{table}

\section{Conclusion}
This work evaluated 1D convolutional neural network architectures and a threshold baseline for gait-event detection in people with Parkinson’s disease using accelerometer data from different body locations. It was shown how chest data clearly outperformed other sensor positions, which included arms and legs. The goal is to detect gait windows on-sensor to replace raw data streams to improve bandwidth, storage, and energy. In leave-one-subject-out validation, all CNNs outperformed the threshold baseline, which in general shows poor generalization. A detailed deployment analysis for embedded systems demonstrated that all models run within memory and latency constraints of representative STM32 microcontrollers. Model 2, which adds residual connections on top of the separable-convolution design of Model 1 and the baseline, achieves the best accuracy–efficiency trade-off, with sub-millisecond inference on H7 and 5--7 ms on L4/F4. Model 1, which employs a simpler separable-convolution architecture, is a valid alternative with slightly lower F1. Fixed-threshold detectors, while extremely cheap, are sensitive to inter-subject variation and under-perform on certain positions relative to CNN models.\\
 \\

\bibliography{aaai2026}

@article{li2025qpart,
  title={QPART: Adaptive Model Quantization and Dynamic Workload Balancing for Accuracy-aware Edge Inference},
  author={Li, Xiangchen and Ghafouri, Saeid and Ji, Bo and Vandierendonck, Hans and John, Deepu and Nikolopoulos, Dimitrios S},
  journal={arXiv preprint arXiv:2506.23934},
  year={2025}
}

@article{dini2024overview,
  title={Overview of AI-models and tools in embedded IIoT applications},
  author={Dini, Pierpaolo and Diana, Lorenzo and Elhanashi, Abdussalam and Saponara, Sergio},
  journal={Electronics},
  volume={13},
  number={12},
  pages={2322},
  year={2024},
  publisher={MDPI}
}

@article{fathalla2022lstm,
  title={An LSTM-based distributed scheme for data transmission reduction of IoT systems},
  author={Fathalla, Ahmed and Li, Kenli and Salah, Ahmad and Mohamed, Marwa F},
  journal={Neurocomputing},
  volume={485},
  pages={166--180},
  year={2022},
  publisher={Elsevier}
}

@article{soumma2024self,
  title={Self-Supervised Learning and Opportunistic Inference for Continuous Monitoring of Freezing of Gait in Parkinson's Disease},
  author={Soumma, Shovito Barua and Mangipudi, Kartik and Peterson, Daniel and Mehta, Shyamal and Ghasemzadeh, Hassan},
  journal={arXiv preprint arXiv:2410.21326},
  year={2024}
}

@article{bikias2021deepfog,
  title={DeepFoG: an IMU-based detection of freezing of gait episodes in Parkinson’s disease patients via deep learning},
  author={Bikias, Thomas and Iakovakis, Dimitrios and Hadjidimitriou, Stelios and Charisis, Vasileios and Hadjileontiadis, Leontios J},
  journal={Frontiers in Robotics and AI},
  volume={8},
  pages={537384},
  year={2021},
  publisher={Frontiers Media SA}
}

@misc{PD-BIOSTAMP-RC21,
  author = {Adams, Jamie L. and Dinesh, Karthik and Snyder, Christopher W. and Xiong, Mulin and Tarolli, Christopher G. and Sharma, Saloni and Dorsey, E. Ray and Sharma, Gaurav},
  title = {{PD-BioStampRC21}: {Parkinson's} Disease Accelerometry Dataset from Five Wearable Sensor Study},
  year = {2020},
  publisher = {IEEE Dataport},
  doi = {10.21227/g2g8-1503},
  url = {https://dx.doi.org/10.21227/g2g8-1503}
}

@article{Borzì-2023,
AUTHOR = {Borzì, Luigi and Sigcha, Luis and Olmo, Gabriella},
TITLE = {Context Recognition Algorithms for Energy-Efficient Freezing-of-Gait Detection in Parkinson’s Disease},
JOURNAL = {Sensors},
VOLUME = {23},
YEAR = {2023},
NUMBER = {9},
ARTICLE-NUMBER = {4426},
URL = {https://www.mdpi.com/1424-8220/23/9/4426},
PubMedID = {37177629},
ISSN = {1424-8220},
DOI = {10.3390/s23094426}
}

@article{chaudhuri2024economic,
  title={Economic Burden of Parkinson’s Disease: A Multinational, Real-World, Cost-of-Illness Study},
  author={Chaudhuri, K Ray and Azulay, Jean-Philippe and Odin, Per and Lindvall, Susanna and Domingos, Josefa and Alobaidi, Ali and Kandukuri, Prasanna L and Chaudhari, Vivek S and Parra, Juan Carlos and Yamazaki, Toru and others},
  journal={Drugs-Real World Outcomes},
  volume={11},
  number={1},
  pages={1--11},
  year={2024},
  publisher={Springer}
}

@article{corra2021comparison,
  title={Comparison of laboratory and daily-life gait speed assessment during ON and OFF states in Parkinson’s disease},
  author={Corr{\`a}, Marta Francisca and Atrsaei, Arash and Sardoreira, Ana and Hansen, Clint and Aminian, Kamiar and Correia, Manuel and Vila-Ch{\~a}, Nuno and Maetzler, Walter and Maia, Lu{\'\i}s},
  journal={Sensors},
  volume={21},
  number={12},
  pages={3974},
  year={2021},
  publisher={MDPI}
}

@article{muthukrishnan2020wearable,
  title={A wearable sensor system to measure step-based gait parameters for parkinson’s disease rehabilitation},
  author={Muthukrishnan, Niveditha and Abbas, James J and Krishnamurthi, Narayanan},
  journal={Sensors},
  volume={20},
  number={22},
  pages={6417},
  year={2020},
  publisher={MDPI}
}

@article{sigcha2020deep,
  title={Deep learning approaches for detecting freezing of gait in Parkinson’s disease patients through on-body acceleration sensors},
  author={Sigcha, Luis and Costa, N{\'e}lson and Pav{\'o}n, Ignacio and Costa, Susana and Arezes, Pedro and L{\'o}pez, Juan Manuel and De Arcas, Guillermo},
  journal={Sensors},
  volume={20},
  number={7},
  pages={1895},
  year={2020},
  publisher={MDPI}
}

@article{mancini2021measuring,
  title={Measuring freezing of gait during daily-life: an open-source, wearable sensors approach},
  author={Mancini, Martina and Shah, Vrutangkumar V and Stuart, Samuel and Curtze, Carolin and Horak, Fay B and Safarpour, Delaram and Nutt, John G},
  journal={Journal of neuroengineering and rehabilitation},
  volume={18},
  pages={1--13},
  year={2021},
  publisher={Springer}
}

@article{kianoush2023random,
  title={A random forest approach to body motion detection: Multisensory fusion and edge processing},
  author={Kianoush, Sanaz and Savazzi, Stefano and Rampa, Vittorio and Costa, Leonardo and Tolochenko, Denis},
  journal={IEEE Sensors Journal},
  volume={23},
  number={4},
  pages={3801--3814},
  year={2023},
  publisher={IEEE}
}

@article{SimonyanZisserman2014,
  title={Very Deep Convolutional Networks for Large-Scale Image Recognition},
  author={Simonyan, Karen and Zisserman, Andrew},
  journal={arXiv:1409.1556},
  year={2014}
}

@inproceedings{HeZhangRenSun2016,
  title={Deep Residual Learning for Image Recognition},
  author={He, Kaiming and Zhang, Xiangyu and Ren, Shaoqing and Sun, Jian},
  booktitle={CVPR},
  year={2016}
}

@inproceedings{Sandler2018,
  title={MobileNetV2: Inverted Residuals and Linear Bottlenecks},
  author={Sandler, Mark and Howard, Andrew and Zhu, Menglong and Zhmoginov, Andrey and Chen, Liang-Chieh},
  booktitle={CVPR},
  year={2018}
}

@inproceedings{LoshchilovHutter2019,
  title={Decoupled Weight Decay Regularization},
  author={Loshchilov, Ilya and Hutter, Frank},
  booktitle={ICLR},
  year={2019}
}

@inproceedings{Cui2019ClassBalanced,
  title={Class-Balanced Loss Based on Effective Number of Samples},
  author={Cui, Yin and Jia, Menglin and Lin, Tsung-Yi and Song, Yang and Belongie, Serge},
  booktitle={CVPR},
  year={2019}
}

\end{document}